\def\year{2021}\relax
\documentclass[letterpaper]{article} % DO NOT CHANGE THIS
\usepackage{aaai}  % DO NOT CHANGE THIS
\nocopyright
\usepackage{times}  % DO NOT CHANGE THIS
\usepackage{helvet} % DO NOT CHANGE THIS
\usepackage{courier}  % DO NOT CHANGE THIS
\usepackage[hyphens]{url}  % DO NOT CHANGE THIS
\usepackage{graphicx} % DO NOT CHANGE THIS
\usepackage{multirow}
\usepackage{threeparttable}
\usepackage{mathrsfs}
\usepackage{amsmath}
\usepackage{booktabs}
\usepackage{dblfloatfix}
\usepackage{booktabs}
\usepackage{amsfonts,amssymb}
\usepackage{array}
\usepackage[switch]{lineno}
\newcolumntype{P}[1]{>{\centering\arraybackslash}p{#1}}
\usepackage{graphicx}  % DO NOT CHANGE THIS
\title{DialogueTRM: Exploring the Intra- and Inter-Modal Emotional Behaviors in the Conversation}
\author{Yuzhao Mao,$^\dag$ Qi Sun,$^\dag$ Guang Liu,$^\dag$ Xiaojie Wang,$^\ddag$ \\ \Large \textbf{Weiguo Gao,$^\dag$ Xuan Li,$^\dag$ \and Jianping Shen$^\dag$}\\ % All authors must be in the same font size and format. Use \Large and \textbf to achieve this result when breaking a line
$^\dag$ PingAn Life Insurance Company of China, Ltd.\\
$^\ddag$ School of Computer Science, Beijing University of Posts and Telecommunications\\ %If you have multiple authors and multiple affiliations
\{maoyuzhao258, sunqi149, liuguang230\}@pingan.com.cn, xjwang@bupt.edu.cn,\\
\{gaoweiguo801, lixuan208, shenjianping324\}@pingan.com.cn % email address must be in roman text type, not monospace or sans serif
}
 
\begin{document}
\maketitle

\begin{abstract}
Emotion Recognition in Conversations (ERC) is essential for building empathetic human-machine systems. Existing studies on ERC primarily focus on summarizing the context information in a conversation, however, ignoring the differentiated emotional behaviors within and across different modalities. Designing appropriate strategies that fit the differentiated multi-modal emotional behaviors can produce more accurate emotional predictions. Thus, we propose the DialogueTransformer to explore the differentiated emotional behaviors from the intra- and inter-modal perspectives. For intra-modal, we construct a novel Hierarchical Transformer that can easily switch between sequential and feed-forward structures according to the differentiated context preference within each modality. For inter-modal, we constitute a novel Multi-Grained Interactive Fusion that applies both neuron- and vector-grained feature interactions to learn the differentiated contributions across all modalities. Experimental results show that DialogueTRM outperforms the state-of-the-art by a significant margin on three benchmark datasets.
\end{abstract}
 
\section{Introduction}
Emotion Recognition in Conversations (ERC) aims to identify the emotions of interlocutors from their textual, visual, and acoustic emotional expressions in conversations~\cite{poria2019emotion}. Techniques in ERC can be widely applied in many fields, e.g., AI interviews, personalized dialogue systems, e-health services, opinion mining systems, etc.
 
Existing studies on ERC primarily focus on summarizing the context information by capturing the self- and inter-personal dependencies in a conversation. Self-dependency deals with the aspect of emotional influence that speakers have on themselves during a conversation~\cite{kuppens2010emotional}. Inter-personal dependency relates to the emotional influences that the counterparts induce into a speaker~\cite{morris2000emotions}. Various models have been proposed to better utilize the two different emotional influences~\cite{jiao-etal-2019-higru,ghosal-etal-2019-dialoguegcn,majumder2019dialoguernn,li2020bieru}.

\begin{figure}[t]
	\centering
	\includegraphics[scale=0.58]{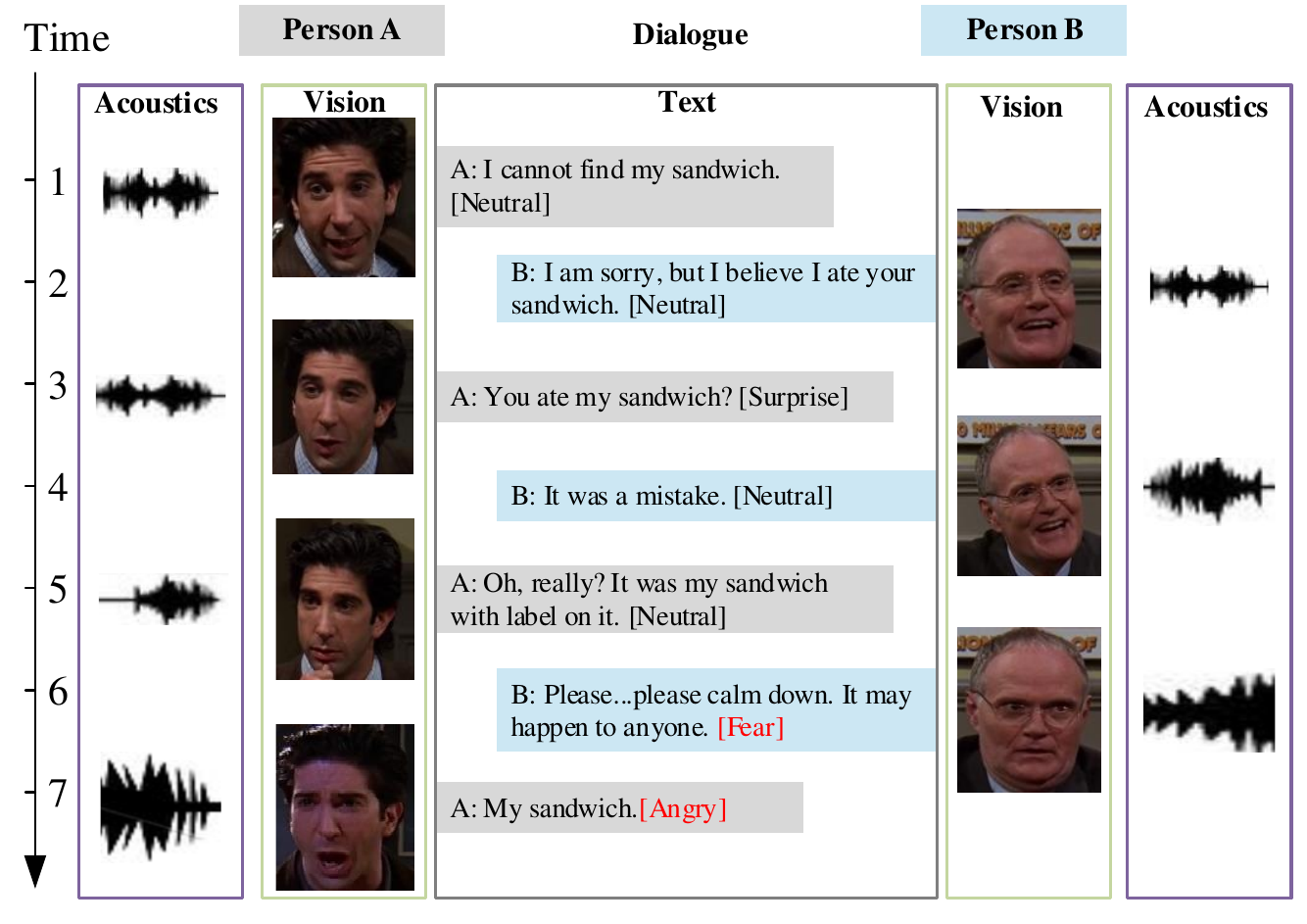}
	\caption{Example of a multi-modal conversation.}
	\label{fig:introduction}
\end{figure}

Despite the progress of previous studies on ERC, few analyses are conducted on the differentiated emotional behaviors within and across different modalities. Notice that language, vision, and speech are three modalities that a machine may process to perceive human emotions. Thus, one specific emotion can be conveyed through the three modalities of emotional expression. However, the point is that the emotional behaviors are not strictly synchronized either within or across modalities when expressing the specific emotion, especially in a conversation. This phenomenon can be interpreted from two aspects as follows: 

One is from the intra-modal aspect that emotional expressions in different modalities have differentiated preferences for conversational context\footnote{In this paper, the notion of context denotes the temporally preceding information of the current expression in a conversation.}. For instance, in Figure \ref{fig:introduction}, let us focus on the temporal dimension within every single modality. Noting from the textual point, it is hard to recognize the ``angry'' emotion based on the only utterance ``My sandwich'' at time 7, while it becomes easy for recognition by looking back to infer that A is angry because B ate his sandwich. From the visual and acoustic point, the emotional tendency of A at time $7$ can be captured instantaneously from the faces or voices. Thus, we argue that textual expressions are cumulatively produced and strongly dependent on preceding utterances. Therefore, context-dependent settings are preferred. On the other hand, visual and acoustic expressions are burst instantaneously so that the context information becomes less useful for recognizing the current emotion. Thus, context-free settings are preferred.

The other is from the inter-modal aspect that emotional expressions in different modalities have differentiated contributions for emotional predictions. It may be observed in Figure \ref{fig:introduction}, if we focus on the spatial dimension across multiple modalities. For person B at time 2, textual and acoustic expressions contribute more to the prediction of ``neutral'' emotion than visual expression (smiling faces often lead to ``happy'' emotion). Thus, to fuse information from multiple modalities, it is essential to learn the contributions or weights of different modalities when predicting emotions. However, representations of different modalities often fall into different spaces, which makes it hard to directly measure the contributions. To tackle this situation, it is essential to first make it comparable between neuron features of different modalities, and then learn the importance of each vector constructed by corresponding neurons. Such multi-grained fusion strategies can better learn the contributions of different modalities.
 
In this paper, we propose a DialogueTRansforMer (DialogueTRM) to explore the intra- and inter-modal emotional behaviors in a conversation. For intra-modal, the temporal dependency within every single modality is the major factor. A Hierarchical Transformer (HT) is constructed by cascading the Transformer~\cite{vaswani2017attention} and the Bidirectional Encoder Representations from Transformers (BERT)~\cite{devlin2019bert}. By applying specially designed attention masks, the HT can switch between sequential and feed-forward structures to manage the context preference within every modality. For inter-modal, the spatial dependency across all modalities is the major factor. A Multi-Grained Interactive Fusion (MGIF) is constituted by stacking the multi-modal gate and the Transformer. The multi-modal gate is designed for neuron-grained fusion that deals with the feature space problem, and the Transformer is employed for vector-grained fusion that learns the contribution of the gated representations. By comprehensively modeling the intra- and inter-modal emotional behaviors in a conversation, our DialogueTRM achieves more accurate emotional predictions than State-Of-The-Art (SOTA) in ERC.
 
In precise terms, our contribution can be summarized as:
\begin{itemize}
\item We explore the intra- and inter-modal emotional behaviors in a conversation for ERC.  Specifically,
      \begin{itemize}
      \item For intra modal temporal modeling, we construct a novel HT module that can manage the differentiated context preference within each modality. 
      \item For inter-modal spatial modeling, we constitute a novel MGIF module that applies multi-modal fusion through both neuron- and vector-grained interactive weighting across all modalities.
      \end{itemize}
\item The proposed DialogueTRM achieves SOTA performance on three ERC benchmark datasets and outperforms SOTA fusion techniques in ERC settings.
\end{itemize}

\section{Related Work}

Emotions are hidden mental states associated with thoughts and feelings~\cite{poria2019emotion}. Without physiological signals, they are only perceivable through human behaviors like textual utterances, visual expressions, and acoustic signals.

\textbf{Emotion recognition} is an interdisciplinary field that spans psychology, cognitive science, machine learning, natural language processing, and others~\cite{picard2010affective}. It involves handling textual, visual, and acoustic sources of data. Early studies on emotion recognition are usually single-modal oriented~\cite{ekman1993facial,schroder2003experimental,strapparava2004wordnet}. Pioneers have explored the advantages of combining facial expressions and speech signals to predict emotions~\cite{tzirakis2017end,wollmer2010context,datcu2014semantic,zeng2007audio}. Recent studies~\cite{tsai2019multimodal,liang2018multimodal,wang2019words,tsai2019learning} have considered all the three modalities, whose primary focus is on fusion strategy while ignoring the emotional dynamics in a conversation. Notice that~\cite{tsai2019learning,liang2018multimodal} also consider the intra-modal and inter-modal interactions, however, the authors ignore the context preference for each modality.

\textbf{Emotion recognition in conversations} is different from traditional emotion recognition. Rather than treating emotions as static states, ERC involves emotional dynamics in a conversation, in which the context plays a vital role. By comparing with the recent proposed ERC approaches~\cite{hazarika2018icon,majumder2019dialoguernn}, ~\cite{poria2019emotion} discovered that traditional emotion recognition approaches ~\cite{colneric2018emotion,kratzwald2018decision,shaheen2014emotion} failed to perform well, because the same utterance within variant context may reveal different emotions. ERC is an advancing topic in the recent few years.~\cite{poria2017context} only captures the self-dependency.~\cite{hazarika2018conversational} and~\cite{hazarika2018icon} distinguish the self and inter-personal dependencies.~\cite{majumder2019dialoguernn} extends ERC to multi-party conversation.~\cite{ghosal-etal-2019-dialoguegcn} uses the Graph Convolutional Network (GCN) to model complex interactions between interlocutors.~\cite{li2020bieru} focus on the party-ignorant transferring of emotion in a conversation. Our model focuses on the emotional behaviors within and across multiple modalities in a conversation.

\textbf{Multi-modal Fusion} seeks to generate a single representation to boost a specific task involving multiple modalities when building classifiers or other predictors. The fusion technics are underestimated in ERC. Multi-modal features are directly concatenated~\cite{hazarika2018icon} or simply ignored~\cite{wang2020contextualized} in previous studies. There are several typical approaches for multi-modal fusion, such as concatenation~\cite{majumder2019dialoguernn}, bilinear~\cite{kiros2014multimodal}, addition~\cite{mao2014deep}, differential operation~\cite{wu2019differential}, gate~\cite{mao2018show}, and attention~\cite{tsai2019multimodal}. Depending on the interactions between features, the above approaches can be categorized into linear weighting fusion (first three) and interactive weighting fusion (latter three).~\cite{gu2019mutual} fuses multiple modalities from the sub-view granularities. Our MGIF considers both the neuron and vector granularities and applies multi-grained fusion.

\section{Task Formulation}
The input in ERC can be defined as a sequence of emotional expressions in $L$ rounds of conversation between $N$ interlocutors, formulated as follows:
\begin{equation}
X\!=\!\{\! x^\lambda_\tau = \!\langle x^\lambda_{\tau,(t)},\!x^\lambda_{\tau,(v)},\!x^\lambda_{\tau,(a)} \rangle\!|\tau\!\in\![ 1, L ],\lambda\!\in\![ 1, N ] \},
\end{equation}
where $x^\lambda_\tau$ is the $\tau$-th emotional expression produced by the $\lambda$-th interlocutor. It comprises three sources of utterance-aligned data in textual ($t$), visual ($v$) and acoustic ($a$) modalities, respectively. Let $\tau=i$, $\lambda=j$, $x_i^j$ be the target emotional expression, $K$ be the context window size. According to whether the interlocutors are distinguished in a conversation, the context is categorized into the individual context 
$\varphi(x_i^j,X,K)\!=\!\{x^\lambda_\tau|\tau\!\in\![i\!-\!K,i),\lambda\!=\!j \}$
and the conversational context 
$\phi(x_i^j,X,K)\!=\!\{x^\lambda_\tau|\tau\!\in\![i-K,i),\lambda\!\in\![1,N]\}$.
Table \ref{tab:example} presents an example of the two types of context. The objective is to predict the emotion of $x^j_i$ given context information in multi-modal settings.

\begin{table}[h]
\centering
\begin{threeparttable}
	\caption{A conversation example, in which context size $K=3$, conversation rounds $L=8$, interlocuter number $N=3$.}
	\small
	\label{tab:example}
	\renewcommand{\arraystretch}{1.5}
	\begin{tabular}{l|l}
		\toprule
		conversation                       & $X=\{x^1_1,x^1_2,x^2_3,x^1_4,x^3_5,x^2_6,x^1_7,x^2_8\}$\\
		\midrule
		target                             & $x^1_7$\\
		\midrule
		individual context                 %& $\varphi(1,7,X,K)=\{x^1_4\}$  \\
                                           & $\varphi(x^1_7,X,K)=\{x^1_4\}$\\
		\midrule
		conversational context             %& $\phi(7,X,K)=\{x^1_4,x^3_5,x^2_6\}$ \\
                                           & $\phi(x^1_7,X,K)=\{x^1_4,x^3_5,x^2_6\}$\\
		\bottomrule
	\end{tabular}
\end{threeparttable}
\end{table}

\section{Model}

The model architecture comprises three modules as depicted in Figure \ref{fig:architecture}. The bottom is the HT module to capture intra-modal emotional behaviors. The middle is the MGIF module to capture inter-modal emotional behaviors. The top is the discriminator module to predict emotions.

\subsection{Hierarchical Transformer}
\begin{figure}[t]
	\centering
	\includegraphics[scale=0.53]{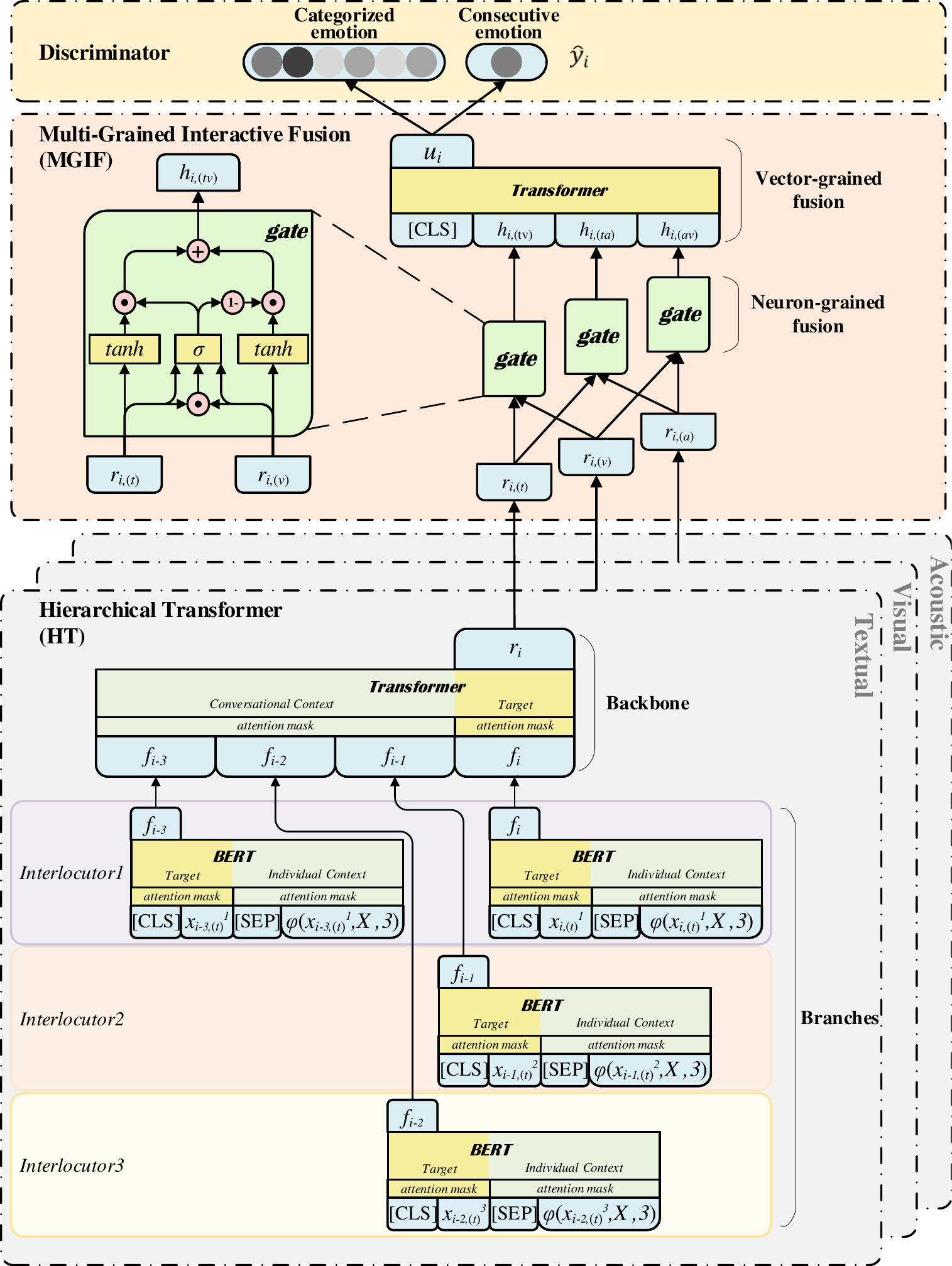}
	\caption{Model architecture. The figure is an example structure by referring Table~\ref{tab:example} where, $\!K\!=\!3$, $N\!=\!3$. In the HT module, the BERT consecutively receives three-modal inputs, and the Transformer wraps all the outputs from BERT. Attention masks manage the context preference by switching the model structure. In the MGIF module, the gate simultaneously receives three-modal representations from HT to make the neuron-grained fusion, and the Transformer wraps the gated representations to make vector-grained fusion.}
	\label{fig:architecture}
\end{figure}

To manage the differentiated context preferences, we construct a novel Hierarchical Transformer (HT) module, shown at the bottom of Figure \ref{fig:architecture}, by cascading the Transformer and the BERT. The structure of HT can be easily switched between a sequential and a feed-forward model through specially designed attention masks in both the Transformer and the BERT. A standard Transformer is composed of a stack of several identical layers. Each layer has a multi-head self-attention mechanism and a simple position-wise fully connected feed-forward network. The BERT is a variant of the Transformer. It inherits most of the advantages of the Transformer and extends particular strategies in dealing with the inputs. The attention mask is a manipulative component in both the Transformer and the BERT that controls whether or not a specific temporal position should be considered in the attention phase.

\textit{Hierarchical Structure.} 
The hierarchical structure is constructed by using the BERT as the branches to handle the inputs and using the Transformer as the backbone to wrap all the information output from the BERT.

The input of the BERT consists of the target emotional expression $x^j_i\!=\!\langle x^j_{i,(t)},\!x^j_{i,(v)},\!x^j_{i,(a)} \rangle$ and the individual context $\varphi(x^j_i,X,K)$. BERT has $3$ key operations in dealing with the pair. 1) Pack the target and context in one sequence;  2) Add $[CLS]$ at the head of a sequence as the classification embedding. 3) Add $[SEP]$ and learnable segment embeddings to differentiate the target and context. The input of the BERT is constituted as $\{[CLS], x^j_i, [SEP], \varphi(x^j_i,X,K)\}$ . The last hidden layer at $[CLS]$ position is used as the output features. The computation of the BERT can be denoted as
\begin{equation}
f^j_i=BERT([CLS], x^j_i, [SEP], \varphi(x^j_i,X,K)),
\end{equation}
where $f^j_i$ is the feature of the target, $i$ and $j$ index the conversation and interlocutors, respectively. Since the next stage is independent of interlocutors, we omit the index $j$.

Let $F=\{f_1,...,f_L\}$ be the features output from the branches in conversational order. The input of the Transformer are the feature of the target $f_i$ and the features of the conversational context $\phi(f_i,F,K)=\{f_{i-K},...,f_{i-1}\}$. The output representation is the last hidden layer at the target position. The computation of the Transformer is
\begin{equation}
r_i=TRM(\phi(f_i,F,K), f_i),
\end{equation}
where $r_i$ is the representation of the target. Since we have three-modal inputs, $r_i$ consists of $r_{i,(t)}$, $r_{i,(v)}$, and $r_{i,(a)}$.

\textit{Context Preference.} 
We adopt two kinds of context preference settings managed by two types of attention masks. One is the context-dependent settings that are suitable for modeling textual expressions. In these settings, the attention mask sets all positions to ones, such that the model forms a sequential structure. Another is the context-free settings that are suitable for modeling expressions in visual and acoustic modalities. In these settings, the attention mask sets the positions of the target to ones and those of the context to zeros, such that the model changes to a feed-forward structure. More precisely, the textual representation hierarchically maintains the self and the inter-personal dependencies from the individual and the conversational context. The visual and acoustic representations preserve instant information within every emotional expression. The three-modal representations with async contextualized information are fed to the MGIF module to learn the spatial dependency across all modalities. 

\textit{Multi-modal encoding.}
The three-modal inputs are first encoded to three-modal features, and then consecutively fed to the embedding layer of BERT. For textual expression, sentences are encoded to a sequence of WordPiece~\cite{devlin2019bert} embeddings learned by pre-trained BERT. For visual expression, videos are encoded by pre-trained 3D-CNN~\cite{tran2015learning}. For acoustic expression, audios are encoded by pre-trained openSMILE~\cite{eyben2010opensmile}. 

\subsection{Multi-Grained Interactive Fusion}
To learn the multi-modal contributions, we constitute a novel Multi-Grained Interactive Fusion (MGIF) module, shown in the middle of Figure \ref{fig:architecture}, by stacking the multi-modal gate and the Transformer. The multi-modal gate is for neuron-grained fusion that adjusts the neuron features of different modalities at the same position. The Transformer is for vector-grained fusion that learns the importance of the gated representations via the built-in attention mechanism.

\textit{Neuron-Grained Fusion.} 
The issue of multi-modal fusion is that the neuron features at the same position are incomparable between different modalities, which causes the vector features of different modalities to fall into different spaces. To make the neuron features comparable, we apply the multi-modal gate~\cite{arevalo2020gated} to allocate contrastive weights to neuron features across different modalities. The weights of neurons are interactively computed and contrastively learned along with the training process. After training, the multi-modal gate can learn the relative importance between neuron feature of different modalities at the same position, so that the feature space problem can be alliviated. Let $r_{i,(t)},r_{i,(v)},r_{i,(a)}\in\mathbb{R}^{D_r}$ be the textual, visual, and acoustic representations at the $i$-th step output from the HT module, the gate operation between textual and visual representations is computed as
\begin{align}
&h_{i,(t)}\!\!\!\!\!\!\!\!\!\!&=&~tanh(W_t \cdot r_{i,(t)})&\\
&h_{i,(v)}\!\!\!\!\!\!\!\!\!\!&=&~tanh(W_v \cdot r_{i,(t)})&\\
&z\!\!\!\!\!\!\!\!\!\!&=&~\sigma(W_z \cdot [r_{i,(t)};r_{i,(v)};r_{i,(t)} * r_{i,(v)}])&\\
&h_{i,(tv)}\!\!\!\!\!\!\!\!\!\!&=&~z*h_{i,(t)}+(1-z)*h_{i,(v)}&
\end{align}
where $h_{i,(t)}, h_{i,(v)}\in\mathbb{R}^{D_h}$ are the projections of $r_{i,(t)}$ and $r_{i,(v)}$. $*$ is referring to the Hadamard product whose function is to use neurons in one vector to weight the neurons of its counterpart at the same position. $\sigma$ is the sigmoid function that maps the weights to $(0,1)$. Our strategy is using neurons in $z$ to weight neurons in $h_{i,(t)}$ and using neurons in $1-z$ to weight neurons in $h_{i,(v)}$, where, $z\in\mathbb{R}^{D_h}$ is computed by feature interactions among $r_{i,(t)}$, $r_{i,(v)}$, and their Hadamard product, $[~;~]$ denotes the concatenation. Note that ``$1-$'' operation performs like softmax in attention mechanism. It normalizes the weights of neurons at the same position, so that the contribution of the neurons can be contrastively learned. The contrastive weighting makes neurons at the same position comparable and additive. $h_{i,(tv)}$ is the bi-modal fused representation.  $W$ and $\cdot$ are the weight matrices and dot product, respectively. We simplify the above computation as
\begin{equation}
  h_{i,(tv)}=GATE(r_{i,(t)},r_{i,(v)})
\end{equation}
Similarly, we can obtain
\begin{equation}
h_{i,(ta)}=GATE(r_{i,(t)},r_{i,(a)}) 
\end{equation}
\begin{equation}
h_{i,(av)}=GATE(r_{i,(a)},r_{i,(v)})
\end{equation}
where $h_i(*)$ represents the neuron-grained fused vectors.

\textit{Vector-Grained Fusion.} 
There are three fused vectors after neuron-grained fusion. We need to learn the contribution of each vector for emotional predictions. In vector-grained fusion, we allocate one weight to an entire vector indicating its importance. Attention~\cite{cho2014learning} is an effective approach for vector-grained interactive weighting. We use another Transformer as the fusion module, where the built-in multi-head attention learns the interactive weights for vector-grained features. Following strategies in BERT, the input is constituted by adding a special embedding $[CLS]$ at the head, which is $\{[CLS],h_{i,(tv)},h_{i,(ta)},h_{i,(av)}\}$. The order of the input is fixed. By feeding the input to the Transformer, the vector-grained fusion can be computed as
\begin{equation}
  u_i=TRM([CLS],h_{i,(tv)},h_{i,(ta)},h_{i,(av)}),
\end{equation}
where $TRM$ is the Transformer. $u_i$ is the output of the last hidden layer at $[CLS]$ position for making predictions.

\subsection{Discriminator}
The discriminator, shown in the top of Figure \ref{fig:architecture}, uses a two-layer perceptron with hidden layer activated by \textit{tanh}. The output can be the softmax for categorized emotion or linear layer for consecutive emotion, formulated as

\begin{equation}
o_i=tanh(W_l \cdot u_i)
\end{equation}
\begin{align}
\mathcal{P}_i&=
\left\{\!
  \begin{aligned}
    &softmax(W_{cat}\!\cdot\!o_i)\!&categorized~emotion\\ 
    &W_{con}\!\cdot\!o_i \!&consecutive~emotion\\
  \end{aligned}
\right.\\
\hat{y}_i&=
\left\{\!
\begin{aligned}
&\mathop{\arg max}\limits_k\mathcal{P}_i[k] &categorized~emotion\\ 
&\mathcal{P}_i                                                  &consecutive~emotion\\
\end{aligned}
\right.
\end{align}
where, $W$ are the weight matrices, $\hat{y_i}$ is the predicted emotion. We use cross-entropy and mean-squared-error loss for categorized and consecutive emotion, respectively,
\begin{equation}
Loss=
\left\{
\begin{aligned}
&\frac{1}{L} \sum^L_i y_i \log( \hat{y_i} ) &categorized~emotion\\ 
&\frac{1}{L} \sum^L_i (y_i - \hat{y_i} )^2 &consecutive~emotion\\
\end{aligned}
\right.
\end{equation}
where, $L$ is the conversation length, $y_i$ is the ground truth.

\begin{table*}[]
\centering
\begin{threeparttable}
	\caption{Categorized emotion results on IEMOCAP and MELD.}
	\small
	\begin{tabular}{p{53pt}|p{15pt}p{15pt}|p{15pt}p{15pt}|p{15pt}p{15pt}|p{15pt}p{15pt}|p{15pt}p{15pt}|p{15pt}p{15pt}|p{15pt}p{15pt}|p{15pt}p{15pt}}
		\toprule
		\multirow{3}{*}{Model} & \multicolumn{14}{c|}{IEMOCAP} & \multicolumn{2}{c}{MELD}\\
		\cline{2-17}
		& \multicolumn{2}{c|}{happy} & \multicolumn{2}{c|}{sad} & \multicolumn{2}{c|}{neutral} & \multicolumn{2}{c|}{angry} & \multicolumn{2}{c|}{excited} & \multicolumn{2}{c|}{frustrated} & \multicolumn{2}{c|}{\textbf{Average}} & \multicolumn{2}{c}{\textbf{Average}}\\
		             & ACC   & F1    & ACC   & F1    & ACC   & F1    & ACC   & F1    & ACC   & F1    & ACC   & F1    & ACC   & F1 & ACC & F1        \\
		\midrule\midrule
		scLSTM       & 37.5  & 43.4  & 67.7  & 69.8  & 64.2  & 55.8  & 61.9  & 61.8  & 51.8  & 59.3  & 61.3  & 60.2  & 59.2  & 59.1& 57.5  & 55.9 \\
		TL-ERC       & -     & -     & -     & -     & -     & -     & -     & -     & -     & -     & -     & -     & -     & 58.8& -        & -    \\	
		%CMN          & 25    & 30.38 & 55.92 & 62.41 & 52.86 & 52.39 & 61.76 & 59.83 & 55.52 & 60.25 & 71.13 & 60.69 & 56.56 & 56.13& -        & -    \\
		%ICON         & 23.6  & 32.8  & 70.6  & 74.4  & 59.9  & 60.6  & \textbf{68.2} & \textbf{68.2} & 72.2  & 68.4  & \textbf{71.9}  & 66.2 & 64.0 & 63.5& -       & -    \\
		DialogueRNN* & 25.69 & 33.18 & 75.1  & 78.8  & 58.59 & 59.21 & 64.71 & 65.28 & \textbf{80.27}& 71.86 & 61.15 & 58.91 & 63.4  & 62.75& 56.1  & 55.9 \\
		DialogueGCN* & 40.62 & 42.75 & \textbf{89.14} & \textbf{84.54} & 61.92 & 63.54 & 67.53 & 64.19 & 65.46 & 63.08 & 64.18 & 66.99 & 65.25 & 64.18& -        & 58.1 \\
		AGHMN        & 48.3  & \textbf{52.1} & 68.3  & 73.3  & 61.6  & 58.4  & 57.5  & 61.9  & 68.1  & 69.7  & 67.1  & 62.3  & 63.5  & 63.5& 59.5  & 57.5 \\
		BiERU*       & \textbf{54.24} & 31.53 & 80.6  & 84.21 & 64.67 & 60.17 & \textbf{67.92} & 65.65 & 62.79 & \textbf{74.07} & 61.93 & 61.27 & 66.11 & 64.65 & -      & 60.84 \\
		\midrule
		DialogueTRM  & 54.19 & 48.7 & 71.04 & 77.52 & \textbf{77.1} & \textbf{74.12} & 64.74 & \textbf{66.27} & 63.91 & 70.24 & \textbf{70.71}  & \textbf{67.23} & \textbf{68.92} & \textbf{69.23} & \textbf{65.66} & \textbf{63.55}\\
		\bottomrule	
	\end{tabular}
	\label{tab:IEMOCAP}
	\begin{tablenotes}
			\scriptsize
			\item Symbol * in both Table~\ref{tab:IEMOCAP} and~\ref{tab:MELD_AVEC} indicate that models are fed with extra succeeding context.
	\end{tablenotes}
\end{threeparttable}
\end{table*}

\section{Experiment}
\subsection{Datasets}
Three datasets are incorporated for ERC, including IEMOCAP ~\cite{busso2008iemocap}, MELD ~\cite{poria2019meld}, and AVEC ~\cite{schuller2012avec}, where the first two orient categorized emotions, and the last orients consecutive emotions.

\textit{IEMOCAP} consists of dyadic conversation videos between pairs of 10 speakers. Each utterance is annotated with one of the six emotional types, including happy, sad, neutral, angry, excited, and frustrated. We follow the previous studies that use the first four sessions for training and use the last session for testing. The validation is randomly extracted conversations from the training set with a ratio of 0.2.

\textit{MELD} consists of multi-party conversation videos collected from Friends TV series. Each utterance is annotated with one of the seven emotional types, including anger, disgust, sadness, joy, neutral, surprise, and fear. MELD provides official splits for training, validation, and testing. The visual source is hard to use unless the speaker can be tracked, which is not the focus of our work. Thus, experiments on MELD do not use visual information. 

\textit{AVEC} consists of dyadic conversation videos between human-agent interactions. It gives every 0.2 seconds four real value annotations to denote the degree of four emotional perceptions from visual and acoustic perspectives, which are Valence, Arousal, Expectancy, and Power~\cite{mehrabian1996pleasure}. We follow strategies~\cite{majumder2019dialoguernn} that use the mean of the continuous values within one emotional expression as the annotation. AVEC provides official splits for training and testing. We make a random extraction from the training set with a ratio of 0.2 for validation. 

\begin{table}[]
\centering
\begin{threeparttable}
	\caption{Consecutive emotion results on AVEC.}
	\small
	\begin{tabular}{p{53pt}|P{32pt}P{32pt}P{32pt}P{32pt}}
	\toprule
                Model & Valence & Arousal& Expectancy& Power \\
		\midrule\midrule
		scLSTM                 & 0.14  &  0.23                  & 0.25   & -0.04 \\
		TL-ERC                 & 0.65  &  0.42                  & 0.35   & -0.03   \\
		%CMN                      & 0.23  &  0.29                  & 0.26   & -0.02   \\
		%ICON                     & 0.237 &  0.297              & 0.26  & 0.225    \\
		DialogueRNN*     & 0.35   & 0.59    & 0.37  & 0.37   \\
		BiERU*           & 0.36   & \textbf{0.64}    & 0.38  & 0.37\\
		\midrule
		DialogueTRM     & \textbf{0.756} & 0.52  & \textbf{0.4}  & \textbf{0.4}  \\
		\bottomrule
	
	\end{tabular}
	\label{tab:MELD_AVEC}
\end{threeparttable}
\end{table}

\subsection{Implementation Details}
The textual features are initialized from BERT and finetuned along with the entire model. The visual and acoustic features are fixed 512- and 100-dimensional vectors, respectively, obtained from the open-source project\footnote{https://github.com/SenticNet/conv-emotion}~\cite{majumder2019dialoguernn}. All features are projected to 768 dimensions to match the input size of the BERT. The BERT is implemented using off-the-shelf pre-trained BERT-base model in ``\textit{Transformers}''\footnote{https://github.com/huggingface/transformers} with default parameters. The Transformer uses a 6-layer, 12-head-attention, and 768-hidden-unit structure implemented by PyTorch using default parameters. Attention masks in both BERT and Transformer are adopted to manage the settings of context preference. The neuron-grained fusion receives three-modal representations of 768 dimensions. The vector-grained fusion adopts a 4-layer, 12-head-attention, and 768-hidden-unit Transformer. We use AdamW~\cite{loshchilov2018fixing} as optimizer with initial learning rate being $6e-6$, $\beta_1=0.9$, $\beta_2=0.999$, L2 weight decay of $0.01$, learning rate warms up over the first $1, 200$ steps, and linear decay of the learning rate. All the results are based on an average of 5 runs. For simplifying and easy-to-reproduce purposes, the proposed model does not expand to multi-GPU settings. Our hardware affords a maximum conversational context size of 14. Larger context can achieve better performance~\cite{hazarika2018icon}, which is beyond the concern of this paper.

\subsection{Baselines and SOTA}
We investigate previous studies tested on the three multi-modal ERC datasets and use them as benchmarks.

\textit{scLSTM~\cite{poria2017context}} is the earliest study that we can track in the task of ERC. It makes predictions considering only the individual context.

\textit{TL-ERC~\cite{hazarika2019emotion}} uses BERT for context modeling. It uses RNN to encode outputs from the BERT and uses another RNN for decoding.

%\textit{CMN~\cite{hazarika2018conversational}} is a memory-network-based model. The results are quoted from ~\cite{hazarika2018icon}.

%\textit{ICON~\cite{hazarika2018icon}} is an extension of CMN. The modification is using another GRU to explicitly model the inter-personal dependency.

\textit{DialogueRNN~\cite{majumder2019dialoguernn}} is a hierarchical attention-based model with two GRUs capturing the context, and another GRU tracking emotional states.

\textit{DialogueGCN~\cite {ghosal-etal-2019-dialoguegcn}} uses GCN to model utterance interactions among interlocutors by considering speaker positions in the historical conversation.

\textit{AGHMN~\cite{jiao2019real}} finetunes sentence representation and uses GRU to wrap the attention-weighted representations rather than summing them up.

\textit{BiERU}~\cite{li2020bieru}(SOTA) applies a party-ignorant bidirectional emotional recurrent unit that fully utilized both sides of the conversational context for emotional predictions

\subsection{Main Results}
We present the categorized emotion results on IEMOCAP and MELD in Table \ref{tab:IEMOCAP} and the consecutive emotions results on AVEC in Table \ref{tab:MELD_AVEC}. We follow~\cite{majumder2019dialoguernn} that uses weighted ACCuracy (ACC) and F1 Score (F1) to evaluate the categorized emotions on IEMOCAP and MELD, and uses pearson’s corRelation coefficient (R) to evaluate the consecutive emotions on AVEC.

For categorized emotion results on IEMOCAP and MELD in Table \ref{tab:IEMOCAP}, it can be noticed that the average scores of our \textit{DialogueTRM} significantly outperform the benchmarks, indicating 4.3\%, 7.1\% improvements on IEMOCAP, 10.4\%, 4.5\% improvements on MELD from SOTA average ACC and F1, respectively. For consecutive emotions results on AVEC in Table \ref{tab:MELD_AVEC}, \textit{DialogueTRM} achieves SOTA performance in most of the criteria, which are 16.3\%, 5.3\%, 8.1\% improvements from SOTA in terms of R for Valence, Expectancy and Power, respectively. The improvements are mainly due to two factors. 1) Characteristic temporal modeling that satisfies the differentiated context preference for each modality. 2) Multi-grained spatial modeling that provides comprehensive multi-modal fusion from different granularities. There are 4 interesting points worth mentioning. Firstly, \textit{DialogueTRM} outperforms \textit{TL-ERC} (using BERT), which reveals the superiority of our model structure rather than simply using BERT. Secondly, \textit{DialogueTRM} outperforms models using succeeding context, including \textit{DialogueRNN}, \textit{DialogueGCN} and \textit{BiERU}, which means our model is more practical in real conversations. Thirdly, from the results of individual emotions in Table \ref{tab:IEMOCAP}, \textit{DialogueTRM} is outstanding in ``neutral''. ``Neutral'' is hard from the textual point because some emotional words, such as the red ``excited'' in Figure~\ref{fig:cases}, could mislead the predictions. \textit{DialogueTRM} can avoid such an issue by referring emotional information from multiple modalities. Fourthly,  The outstanding performance of R score on ``Valence'' emotion in Table~\ref{tab:MELD_AVEC} benefits from BERT features, as \textit{TL-ERC} using BERT also achieves outstanding performance.

\subsection{Inter-modal Results}
We present the fusion results of F1 in Table~\ref{tab:MM} on IEMOCAP. The point is the uplift from text-only to multi-modal which is denoted as subscripts in the multi-modal column. The ``$67.67$'' result in the text-only column is based on our intra-modal module fed with text-only features.

\begin{table}[]
\centering
\begin{threeparttable}
\caption{Fusion results on IEMOCAP.}
\small
\begin{tabular}{p{75pt}|p{55pt}p{65pt}}
\toprule
Model          & Text-only                       & Multi-modal\\
\midrule
dialogueRNN & 62.7~~~~~~~~~~~~$\to$                    & $62.9_{0.2\uparrow}$ \\
\midrule
MulT$\dag$  & \multirow{2}{*}{67.67~~~~~~~~~~$\to$} & $67.98_{0.31\uparrow}$ \\
MGIF$\dag$  &                                 & $68.52_{0.85\uparrow}$ \\
\midrule
\midrule
Gate+TRM (MGIF)   & \multirow{3}{*}{67.67~~~~~~~~~~$\to$}                                & $69.23_{1.56\uparrow}$ \\
Gate+Concat &                               & $68.72_{1.05\uparrow}$ \\ 
TRM &                        & $68.61_{0.94\uparrow}$ \\ 
                         
\bottomrule   
\end{tabular}
\label{tab:MM}
\begin{tablenotes}
		\scriptsize
		\item Symbol $\dag$ denotes all modalities use context-dependent settings.
	\end{tablenotes}
\end{threeparttable}
\end{table}

\textit{Comparison with SOTA.}
\textit{DialogueRNN} is the SOTA model for ERC in multi-modal settings. It uses concatenation to fuse multi-modal information. The uplift of MGIF is seven times higher than that of \textit{DialogueRNN}. The significance benefits from two factors. One is from the interactive weighting. The other is from the multi-grained fusion. MulT~\cite{tsai2019multimodal} is the SOTA fusion technique for emotion recognition. It uses a Transformer to temporally fuse the multi-modal features in an encoder-decoder fashion. Thus, all modalities must use context-dependent settings. For a fair comparison, both MGIF$\dag$ and MulT$\dag$\footnote{https://github.com/yaohungt/Multimodal-Transformer} are fed with context-dependent multi-modal features output from our the HT module. The uplift of MGIF$\dag$ is nearly tripled than that of MulT$\dag$. The superiority of MGIF benefits from interactions between features of varied granularities.

\textit{Ablation study.}
The bottom three rows in Table~\ref{tab:MM} are the ablation results of MGIF module. The results are based on three-modal representations with preferred context information output from the HT module. TRM is short for Transformer. 1) \textit{Gate+TRM} vs. \textit{TRM} is a comparison of feeding the Transformer with or without gated features. It indicates the necessity for neuron-grained fusion. 2) \textit{Gate+TRM} vs. \textit{Gate+Concat} is a comparison between the linear and interactive weighting for fusion. It indicates that interactive weighting (TRM) is better than linear weighting (Concat). 3) The superiority of \textit{Gate+TRM} over the others indicates that the multi-grained fusion is better than single-grained fusion.

\begin{figure}[t]
	\centering
	\includegraphics[width=0.45\textwidth]{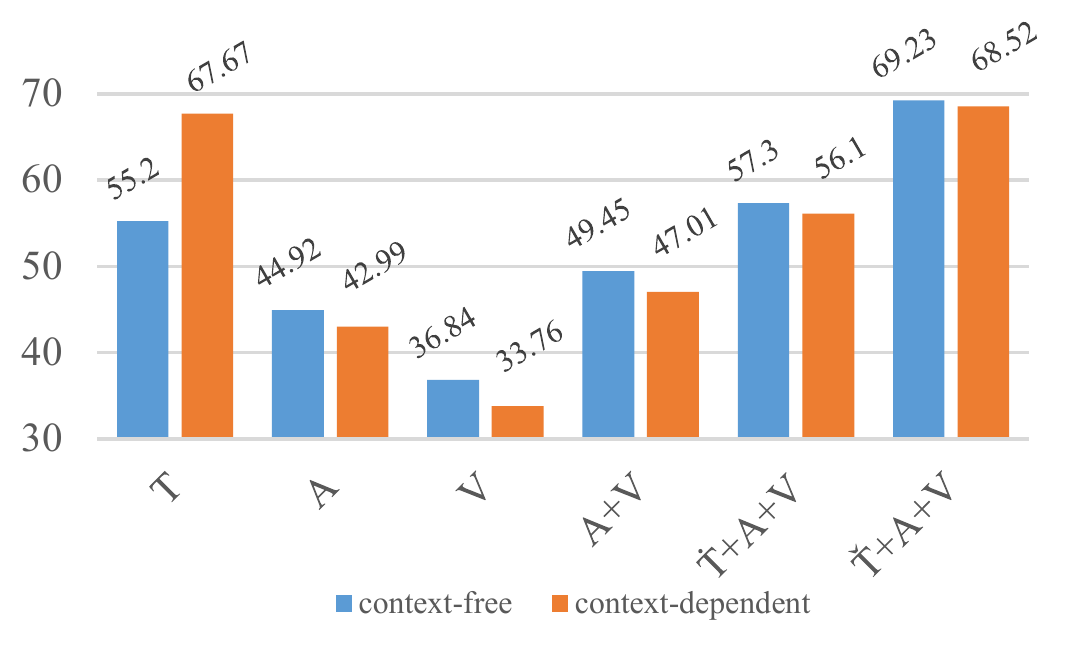}
	\caption{Context perferences results on IEMOCAP. $T$, $A$ and $V$ denote textual, acoustic and visual feature, respectively. $\dot{T}$ and $\check{T}$ respectively denote keeping textual feature context-free and context-dependent.}
	\label{fig:context}
\end{figure}

\subsection{Intra-modal Results}
In this experiment, we first analyze the context preference within each modality and then discuss the different structures for summarizing the textual context. All results are based on the F1 score on IEMOCAP.
 
\textit{Context Preference.}
Due to the different context preference settings, the uplift of MGIF is nearly doubled than that of MGIF$\dagger$ as presented in Table~\ref{tab:MM}. To detail the analysis for context preference, we depict the comparative performance of our model between context-free and context-dependent settings in Figure~\ref{fig:context}. There are 2 points worth mentioning. 1) Textual modality performs much better in context-dependent settings. 2) Visual and acoustic modalities always achieve better performance in context-free settings whether as an individual or after fusion. Note that, to focus on the context preference for visual and acoustic modalities, the context preference settings of textual modality are fixed. 

\begin{figure}[h]
	\centering
	\includegraphics[width=0.45\textwidth]{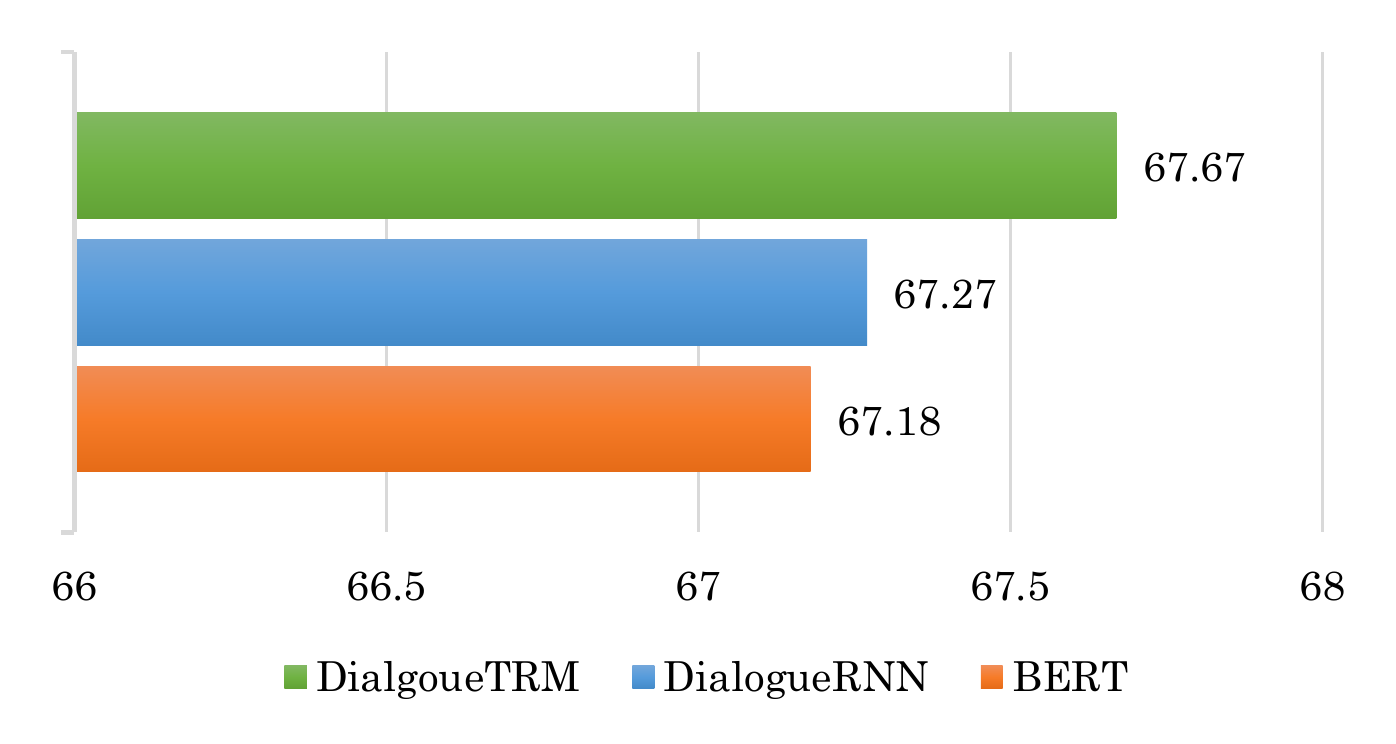}
	\caption{Temporal module result on IEMOCAP.}
	\label{fig:BERT}
\end{figure} 

\textit{Temporal module.}
From the results in Figure~\ref{fig:context}, we notice that the textual modality dominates the performance, and the performance is strongly affected by the context information. In this experiment, we present the performance of different structures for summarizing the textual context. \textit{BERT} makes predictions by summarizing individual context. \textit{DialogueRNN} uses three GRUs with global and local attention to interactively maintain three states in a conversation. \textit{DialogueTRM} uses just the Transformer. For fairness, \textit{DialogueRNN} also uses contextualized features output from \textit{BERT}. The comparison is depicted in figure~\ref{fig:BERT}. Notice that \textit{DialogueTRM} performs better than the others. \textit{DialogueRNN} is slightly better than \textit{BERT} which means it can hardly extract more useful information from \textit{BERT} features.

\begin{figure}[t]
	\centering
	\includegraphics[width=0.5\textwidth]{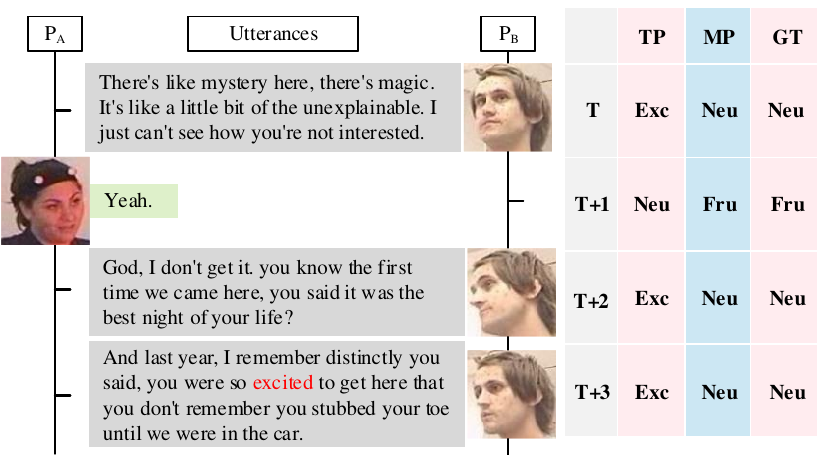}
	\caption{Conversation cases with MP (Multi-modal-Predicted), TP (Text-Predicted) and GT (Ground-Truth) emotions, where 'Neu', 'Exc', 'Fru' stands for neutral, excited and frustrated, respectively. }
	\label{fig:cases}
\end{figure}

\subsection{Case Study}
1) We analyze cases that are incorrectly predicted from the text but correctly predicted from the multi-modal. Among the cases, ``neutral'' and ``frustrated'' are in the majority with the ratios of about 30.38\% and 27.85\%, respectively. It means multi-modal contributes more to ``neutral'' and ``frustrated'' cases. Moreover, about 85.41\% ``neutral'' and 70.45\% ``frustrated'' cases are rectified from negative emotions, which means multi-modal provides easy-to-distinguish information for negative emotions. The reason is probably that human tends to use neutral words to cover their negative emotions but still showing up in the face or speech. 2) Figure \ref{fig:cases} lists a conversation case. The incorrectly text-predicted ``excited'' emotion at time $t+3$ is probably because of the misleading word ``excited'' in the utterance while rectified by multi-modal. 3) The utterance ``yeah.'' appears 23 times in the test set. Given only the current utterance, the accuracy is ``43.48\%''. After adding context, it increases to ``65.22\%''. After adding multi-modal, it reaches ``73.91\%''.

\section{Conclusion}
This paper proposes the DialogueTRM with the HT and MGIF modules for respectively capturing the intra- and inter-modal emotional behaviors in a conversation. It achieves SOTA performance on ERC datasets. Besides, the MGIF module outperforms the SOTA fusion approaches in ERC settings. In the experiment, we analyze the context preferences of emotional expression in different modalities and discuss several cases to reveal the benefit of our model. In the future, we will further explore the emotional behaviors in conversation. It is also possible to incorporate the MGIF inside of the Transformer for constructing a more advanced temporal module.

\bibliographystyle{aaai} 
\bibliography{ref}

\end{document}